# Don't get Lost in Negation: An Effective Negation Handled Dialogue Acts Prediction Algorithm for Twitter Customer Service Conversations

Mansurul Bhuiyan
IBM-Research, Almaden
San Jose, CA, USA
mansurul.bhuiyan@ibm.com

Amita Misra
UC Santa Cruz
Santa Cruz, CA, USA
amisra2@ucsc.edu

Saurabh Tripathy
IBM-Research, Almaden
San Jose, CA, USA
Saurabh.Tripathy2@ibm.com

Jalal Mahmud
IBM-Research, Almaden
San Jose, CA, USA
jumahmud@us.ibm.com

Rama Akkiraju
IBM-Research, Almaden
San Jose, CA, USA
akkiraju@us.ibm.com

## ABSTRACT

In the last several years, Twitter is being adopted by the companies as an alternative platform to interact with the customers to address their concerns. With the abundance of such unconventional conversation resources, push for developing effective virtual agents is more than ever. To address this challenge, a better understanding of such customer service conversations is required. Lately, there have been several works proposing a novel taxonomy for fine-grained dialogue acts as well as develop algorithms for automatic detection of these acts. The outcomes of these works are providing stepping stones for the ultimate goal of building efficient and effective virtual agents. But none of these works consider handling the notion of negation into the proposed algorithms. In this work, we developed an SVM-based dialogue acts prediction algorithm for Twitter customer service conversations where negation handling is an integral part of the end-to-end solution. For negation handling, we propose several efficient heuristics as well as adopt recent state-of-art third party machine learning based solutions. Empirically we show model's performance gain while handling negation compared to when we don't. Our experiments show that for the informal text such as tweets, the heuristic-based approach is more effective.

## INTRODUCTION

In recent years, with the increasing involvement of social media in our everyday lives, online messaging has become a new conversational medium. Among all the different types of conversation that we engage in, interaction with customer care personnel is a common and important one. Nowadays, Twitter has evolved as a prominent channel for customer care conversations, where customers can make inquiries and receive an instant reply by the Twitter representative of the corresponding company. For a customer care representative, it is very important to understand the essentials of the conversation (Twitter or other forms) in order to construct an informative and effective reply. An effective understanding of conversations can bring tremendous value to better realize the requirements of the conversation for human agents, and can also fuel the development of intelligent virtual agents.

One of the important insights regarding a conversation is its flow and content. Dialogue act is one of the established methods to model the interactions in a conversation and describe its flow. Each turn of a conversation can be associated with different dialogue acts to portray its semantic meaning at a higher level. For example, `Why my phone is not working?` can be labeled by a dialogue act of type Wh-question (Question starts with why, where, when etc) and a complaining statement. This information tells us that the next consecutive turn will be an answer of some sort. The study of dialogue acts can be leveraged to measure the quality of a conversation by analyzing what sequences (dialogue acts) lead to a successful interactions. Recently in [20], authors have proposed a new taxonomy of 25 dialogue acts tailored to twitter based customer service conversations. They also show how automatic detection of dialogue acts can be useful for the development of virtual agents.

Another important aspect for an effective understanding of a conversation comes from the linguistic point of view. In the above example, we can see an existence of a negation cue "not" that is essentially inverting the meaning of the word "working". To elaborate, cue words like "no", "not", "never" and "don't" are used to negate a statement or an assertion that expresses a judgment or an opinion [15]. For virtual agents, it is of utmost important to distinguish between "working" and "not working". In this work, we develop a dialogue-acts prediction algorithm that takes negation handling into consideration. To our knowledge, we are the first to address negation handling for dialogue acts prediction in twitter based customer care conversations.

Our contributions are following: We collect 2235 customer care utterances containing negations (one or more negation cues) from twitter for different companies. We use the same ontology as [20], where we pick top 12 acts (same set used by the authors during empirical evaluation of their system) and use crowd-annotators to collect ground truth label. We develop several time efficient heuristics for negation detection and compared with a recent CRF-based approach. We design an utterance level dialogue acts prediction algorithm using SVM for twitter-based customer care conversations. In the following sections, we will discuss details of the negation handling approaches, feature engineering and modeling of dialogue acts prediction algorithm and empirical evaluations.

## RELATED WORK

In this section we will be discussing related works for dialogue acts prediction and negation scope detection.

### Dialogue acts

In this section, we will discuss existing works on dialogue acts modeling. Several works explored speech act modeling which is the previous version of dialogue acts modeling on Twitter [31]. In these works, authors predict five acts *Statement, Question, Suggestion, Comment*, and a *Miscellaneous* from a tweet. As a followup, Vosoughi et al. developed [28] a classification algorithm for speech acts detection on Twitter using an updated taxonomy of six acts *Assertion, Recommendation Expression, Question, Request*, again plus a *Miscellaneous* tailored to Twitter. In this work, we are more interested in detecting dialogue acts in a tweet that is part of a conversation between a customer and an agent.

In 1997 [2], Core and Allen present the Dialogue Act Marking in Several Layers (DAMSL). The taxonomy was made of 220 tags distributed into four categories: communicative status, information level, forward-looking function, and backward-looking function. Later on Jurafsky et. al [9] developed a less fine-grained taxonomy of 42 tags based on DAMSL. Over the years, there has been several works on developing dialogue acts taxonomies [1, 14, 25, 27, 20] for general conversations or customer service conversations. In this work, we leverage the taxonomy proposed by [20].

In 2005, Ivanovic explored interactions between customers and agents in instant messaging chats [7, 8] using a set of of 12 course-grained dialogue acts. Afterwards Kim et al. focus on classifying dialogue acts in both one-on-one and multi-party live instant messaging chats [10, 11]. In 2017, Soraby et.al [20] proposed a set of 25 dialogue acts for customer service conversation on Twitter and developed a sequential prediction algorithm. Our work is similar to the [20]'s task but the main difference is that we study the effects of the negation in each utterance levels of a conversation and develop a non-sequential dialogue acts prediction algorithm.

### Negation scope detection

Initial explorations on negation scope detection were performed mostly in Biomedical domain where it is important to distinguish facts from negative assertions [26] in medical reports, biological abstracts and full papers. [18] presents a two steps approach: first, a decision tree to predict negation cues, followed by a CRF metalearner to predict the scope of negations that combines input from k-nearest neighbor classification, a support vector machine, and another underlying CRF.

A rule-based and a data driven approach is also explored [22] for scope resolution. The rule-based approach uses POS tags and constituent category labels based heuristic rules. On the other hand, machine learning uses SVM based ranking of syntactic constituents. [3, 16, 23, 24] proposed CRF-based sequence labeling leveraging features from dependency tree. [21] uses hand-crafted heuristics to traverse Minimal Recursion Semantics (MRS). [5] show that a neural network based model using Bidirectional Long Short Term Memory, (BiLSTM) outperformed the previously developed classifiers on both scope token recognition and exact scope matching for in domain testing but not on a different domain. The authors note that when tested on a different test set from Simple Wikipedia, [29]'s model built on constituency-based features performed better. In this work, to compare we pick the most recent CRF based model proposed by [24], we could not compare with [5] because the model expects testing data to be processed and manually annotated using the guideline from [19] and our conversation data is not formatted as per [19].

## NEGATION MODELING

In this section, we discuss the heuristics and methods used to handle negation for a given tweet. In general, negation detection has two parts: negation cue detection and scope detection. Majority of the proposed methods use proper written English text but in this paper we tackle informal text like tweets. Also, none of the works address dialogue acts prediction in the negation setting.

| hardly | lack | lacking | lacks | neither |
|---|---|---|---|---|
| no | nobody | none | nothing | nowhere |
| cant | arent | dont | doesnt | didnt |
| havent | isnt | mightnt | mustnt | neednt |
| shouldnt | wasnt | werent | wouldnt | without |
| seldom | scarcely | wont | never | aint |
| barely | nor | not | hadnt | rather |
| hasnt | shant | | | |

**Table 1: Negation Cue Lexicon**

### Negation Cues

In Natural language, a negation can appear in both syntactic as well as pragmatic stages. The existence of syntactic negation is possible via the use of explicit negation cues such as *no, not, never*. These explicit negation cues have a direct effect on the following expression and can change its meaning. There can also be some semi-negative words i.e. rarely, seldom as an instance of explicit negation. Negation can also appear via affixes such as, *in-, im-, dis* or suffixes *-less* and *-out* but theirs implication is limited to the single word. As we are interested in how negation affects these expressions, in this work we only focus on explicit negation listed in [3]. Table 1 presents the explicit negation cues.

### Negation Scope Detection

In this section, we will discuss different scope detection algorithms.

#### Parts-of-Speech (POS) Based

The proposed parts-of-speech (POS) based negation scope detection is working as follows:

**Data:** tweet
**Result:** Identified Scope
identify the negation cue;
**while** *next word after cue* **do**
　**if** *word is a stop word* **then**
　　| continue;
　**end**
　**if** *word is in* [, . :; ?!] **then**
　　| return "no scope found hit punctuation";
　**end**
　**if** *word is another negation cue* **then**
　　| return "no scope found hit another cue";
　**end**
　**if** *POS of word is in* ["NN","JJ","VB"] **then**
　　| return word;
　**end**
**end**
**Algorithm 1:** Parts-of-Speech based negation scope detection

For an input like "I am not at all happy with your service.", The POS-based approach will first identify the negation cue which is *not*. After that, it starts scanning for the next available word which is not a punctuation or a stop-word or another cue and has POS of either noun, adjective or verb (excluding auxiliary verb). In this example, *at* and *all* will be skipped and *happy* will be returned as the scope. The idea of POS-based negation scope detection is that we want to identify scope that is the minimum of length and first available noun, adjective or verb will be the word that is mostly affected by the negation cue.

*First Emotional word*

This detection algorithm is similar to POS based approach but the difference is it looks for first available emotional word after the negation cue and return as scope. To identify emotional words i.e. bad, good, anxious, we leverage five publicly available emotional and sentiment dictionaries. Details of these dictionaries are discussed in Section .

*Window based*

In this heuristic, after identifying the negation cue, we indicate X words before and Y words forward of the cue word as scope. If we set X to 1 and Y to 2 for an input "I am definitely not very happy with this service", then according to this heuristic "definitely" (1 word before) and "very happy" (2 words after) constitutes the scope. Authors of [6] proposed similar heuristic but for fixed size window before and after. In our experiments, we vary X and Y from 0 to 3. In case of multiple negation cues two things can happen: i) another negation cue can be in the window span of the current cue in hand ii) the span of windows for each cue might get overlapped. To handle the first case, we only keep words from the window that are not negation cues. For the second case, we keep track of the words already being selected to be in the scope of a cue and do not consider them anymore for the next cue scope resolution.

*Punctuation Based*

This heuristic is used as a well-known baseline in existing negation scope detection literature [13]. According to this heuristic, all the words after a cue until a punctuation is considered to be in scope. In this work we use {, . :; ?!} as punctuations.

*Conditional Random Field Based*

We also use a recent Conditional Random Field (CRF) based machine learning algorithm [24] for negation scope detection. In [24], authors trained a CRF model by utilizing several dependency parse tree based features: i) POS tag of the 1st order dependency head, ii) number of edges to nearest negation cue from the 1st order dependency head iii) Same as above two for 2nd order dependency head iv) distance to nearest negation cue to the right and left v) word and its POS. As we could not get the source code or executable from authors, we implemented the algorithm by our-self in Python.

## DATA

data In this section, we will discuss the data annotation guidelines, report agreement, and basic data statistics.

## Annotation Guideline

As discussed earlier we pick top 12 dialogue acts proposed in [20]. As our main focus is to study the effect of negation, during our data collection we only select customer care tweets from various companies with one or more negation cues from Table 1. In total, we collect 2235 utterances of customer care tweets. During annotation, we show one utterance at a time to the annotator and ask to pick dialogue acts from the list he/she thinks appropriate. Each tweet is annotated by 5 annotators and we finalize label by taking a majority vote. We use crowdflower.com for crowd annotation. As we could not maintain a list of common annotators across the task, as agreement score we compute an average of the percentage of annotators agree on a label given a tweet. We found 73.3% annotators agreement in the data.

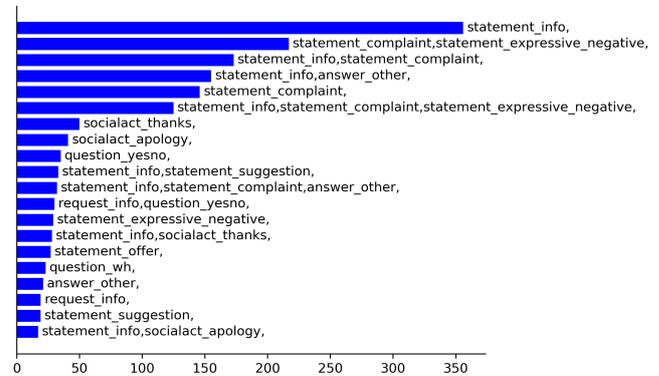

**Figure 1: Crowd annotated label (Dialogue Acts) distribution**

## Data Statistics

We present two types of data statistics. First, we compute the distribution of negation cues in the data. We found almost 83% (1863 samples) of the data contains a single negation cue. The number of instances with 2,3 and 4 negation cues are 325, 40 and 7 respectively. There is no sample with more than 4 negation cues in the data. Next, we compute annotated dialogue acts distribution and we found out that a large number of samples have multiple annotated dialogue acts. To be precise we found 191 different combinations of annotated dialogue acts. In Figure 1 we show the distribution of top 20 label combinations.

## METHOD

In this section, we describe the methodology of predicting dialogue acts in customer service conversation while incorporating negation detection module in the pipeline.

## Data Pre-processing

We have incorporated following pre-processing steps into the pipeline.

*Cleaning*

In this step, we do following cleanups: i) Replace all website urls (https://www.twitter.com) by a placeholder word "URL". ii) Remove # from the hashtags. iii) Replace all @$mentions$ by a keyword "ATUSER" iv) Replace emojis and emoticons with the word explanation. i.e. :) becomes "happy" v) Consolidate additional white spaces. vi) Expand abbreviation to its full form i.e. brb becomes be right back.

*Entity Resolution*

In this stage, we run the entire dataset with internally developed entity detection service and identify five types of entity: person's name, country, county, city, and company. Then we manually check the list to resolve any confusion, for example, "joy" can be a person's name and "joy" is also a very important emotional word. Finally, we replace all the filtered identified entity in the data using a

keyword "ENT". The motivation behind this cleaning procedure is to reduced sparsity when we generate word-based features.

### Negation handling

In this module, we integrated the developed negation scope detection algorithm(s) into the prediction pipeline. For each cleaned tweet we execute this module and received a transformed version of the input after handling negation. The way we are transforming the input text is following. After detecting the scope we modify each word in the scope by pre-pending the word "NOT_" and remove associated negation cue. For example, `I am not :) with the service` is the original text, after clean up it becomes, `I am not happy with the service`. Now suppose we detect "happy" (emotion and POS based method which detect happy as scope) to be in the scope of negation cue "not". After applying transformation it returns `I am NOT_happy with the service`. Such transformation will allow model to distinguish between `I am :) with the service` and `I am not :) with the service` while generating features specially n-grams for the model to use.

### Features

In this section, we will explain the set of features used in the proposed dialogue acts prediction algorithm.

**Word based:** We use TFIDF-based unigram features. In the TFIDF dictionary, we set minimum term frequency count of a word to 2. We handle the result of negation detection via this feature. For the above example, if we do not handle negation, in the dictionary there will be a single entry related to "happy". Whereas with negation handling there will be "happy" and "NOT_happy" as dictionary entries.

**Emotion lexicon features:** We count the number of words in each of the 8 emotion classes from the NRC emotion lexicon [17] (anger, anticipation, disgust, fear, joy, sadness, surprise, and trust).

**Sentiment lexicon features:** We used the following five sentiment lexicons to derive additional set of features.

i) *Bing Liu's Opinion Lexicon* [4]. It contains a set of 2006 positive and 4783 negative sentiment words that have a score of -1 for negative and +1 for positive words.

ii) *The MPQA Subjectivity Lexicon* [30]. It contains a list of 8222 English words with scores ranging from -2 to +2 indicating intensity of negative and positive sentiment.

iii) *Sentiment140 Lexicon* [13]. This lexicon contains a real valued sentiment score of English words as well as hashtags. The scores are computed while considering negated contexts. For example, "happy" can have two different scores depending on if the word is being used in a negated context or not. Authors of the lexicon using "NEG" and "NEGFIRST" tag to indicate a word being used in a negated context. The details on how the scores are computed can be found in [13].

iv) *NRC Hashtag Sentiment Lexicon* [13]. This is same as the Sentiment140 Lexicon, but only for hashtags.

v) *NRC Emotion Lexicon* [17]. This lexicon has a binary value of positive and negative sentiment for English words.

For a given tweet, using all the lexicons mentioned above we computed minimum, maximum, average and summation of positive and negative scores of the words in the tweet that lies within a negation scope. For example, if a scope contains 5 words and out-of 5, 3 are positive and 2 are negative, then we measure above four numerical measures of 3 and 2 words respectively. We do the same for the words that are outside of negation scope. As an additional feature, we also compute average sentiment score of a last word in the tweets using all the dictionaries.

**High level features:** We also computed following higher level features that we think have potential of better sentiment prediction in customer service domain:

i) Existence of consecutive question and exclamation marks. In customer service domain people often used consecutive question or exclamation marks to portray frustration or excitement respectively.

ii) Existence of single question mark. People often ask questions while complaining.

iii) Number of capitalized word. When frustrated, customers often used all caps writing style.

**Dialogue:** We used several dialogue related lexicons to derive indicator features on i) yes-no questions (questions beginning with did, do, can, could, etc) ii) wh- questions (turns with questions starting with who, what, where, etc) iii) opening greetings (hi, hello, greetings, etc) iv) closing greetings (bye, goodbye) and v) thanking and apology (sorry, thanks, thank you etc)

### Model

To build the dialogue acts classification model we use SVM with linear kernel. We did not use current state-of-the art CNN based text classification models [12] due to limited amount of training data. We tried random-forest and Logistic regression model but SVM with linear kernel was the best performing one. As we seen in Figure 1, that multiple dialogue acts can be associated to an input text and for that we are training one binary classifier for each of the dialogues acts (class) using One-VS-Rest paradigm. During prediction, we select all classes that have a probability 0.5 or higher.

## EXPERIMENT

As we want to establish the benefit of integrating negation scope detection in the dialogue acts prediction, in the first experiment we compare classification accuracy (FScore) between two different setups of the execution: i) we do not include negation detection ii) we perform negation detection using all the heuristics presented in the negation modeling Section. We use 5 fold cross-validation, 3 folds for training and 1 fold for tuning model's parameter and 1 fold for testing.

### Dialogue Acts Prediction With and Without Negation

We perform this experiment into two parts. First, we empirically find out which window based heuristic performs best for the majority of dialogue acts prediction. In Table 2 we present the result. We found out that, among all the variations -1/+2 heuristic performs best. This heuristic declares one previous word and two next as a scope of a given negation cue. From this point onward we select -1/+2 window based heuristics to compare with others.

In the second part of this experiment, we compare when we do not include any negation scope detection vs when we do. In Table 3, we present the findings. As we can see, for all the acts except *socialact-thanks* there is a performance improvement when we do some form of negation detection and handling. One possible reason for *socialact-thanks* act can be, this act is quite easy to detect (Higher FScore supports that). As most of the time, such thanking statement contains strong and clear words indicating "thank you" apart from negation cue(s). Among different heuristics, even though there is no clear winner but majority wise window based approach performs best. One possible reason CRF method did not perform well is that, the model is trained on Sem-Eval 2012 Shared task on negation scope detection data[1], which is derived from Arthur Conan Doyal's story (formal English) and tested on informal text like tweets that most often do not follow proper grammar.

---
[1] http://www.clips.ua.ac.be/sem2012-st-neg/data.html

| Dacts | -1/+1 | -2/+2 | -3/+3 | -2/+3 | -1/3 | 0/3 | -1/+2 | 0/2 |
|---|---|---|---|---|---|---|---|---|
| statement_info | 0.66 | 0.68 | 0.69 | 0.67 | 0.68 | 0.69 | **0.7** | 0.69 |
| request_info | 0.61 | 0.58 | 0.57 | 0.59 | 0.61 | 0.6 | **0.62** | 0.59 |
| statement_complaint | 0.65 | 0.66 | 0.67 | 0.66 | 0.64 | 0.66 | **0.67** | 0.65 |
| question_yesno | 0.5 | 0.43 | 0.37 | 0.42 | 0.46 | 0.46 | **0.46** | 0.45 |
| statement_expressive_negative | 0.52 | 0.49 | 0.53 | 0.52 | 0.53 | 0.5 | **0.53** | 0.5 |
| statement_suggestion | 0.44 | 0.32 | 0.31 | 0.32 | 0.32 | 0.35 | **0.38** | 0.36 |
| answer_other | 0.28 | 0.24 | 0.21 | 0.26 | 0.22 | 0.3 | **0.34** | 0.31 |
| socialact_thanks | 0.91 | 0.85 | 0.8 | 0.82 | 0.87 | 0.87 | **0.89** | 0.89 |
| question_wh | 0.7 | 0.71 | 0.61 | 0.62 | 0.62 | 0.63 | **0.72** | 0.57 |
| statement_offer | 0.46 | 0.52 | 0.5 | **0.54** | 0.41 | 0.44 | 0.41 | 0.46 |
| question_open | 0.33 | 0.36 | 0.25 | 0.29 | 0.34 | 0.35 | **0.36** | 0.35 |
| socialact_apology | 0.93 | 0.92 | 0.91 | 0.88 | 0.88 | 0.88 | **0.93** | 0.93 |

Table 2: Performance of Window-based approach

| Dacts | without | Window | POS | Punctuation | Emotional | CRF |
|---|---|---|---|---|---|---|
| statement_info | 0.68 | **0.7** | 0.69 | 0.68 | 0.68 | 0.68 |
| request_info | 0.62 | **0.62** | 0.62 | 0.59 | 0.62 | 0.62 |
| statement_complaint | 0.64 | 0.67 | 0.62 | **0.68** | 0.65 | 0.63 |
| question_yesno | 0.39 | 0.45 | 0.48 | **0.49** | 0.44 | 0.45 |
| statement_expressive_negative | 0.48 | **0.53** | 0.49 | 0.48 | 0.44 | 0.42 |
| statement_suggestion | 0.47 | 0.38 | **0.48** | 0.34 | 0.45 | 0.4 |
| answer_other | 0.21 | **0.34** | 0.23 | 0.29 | 0.26 | 0.3 |
| socialact_thanks | **0.93** | 0.89 | 0.8 | 0.5 | 0.8 | 0.78 |
| question_wh | 0.64 | **0.72** | 0.6 | 0.46 | 0.59 | 0.54 |
| statement_offer | 0.44 | 0.41 | **0.45** | 0.43 | 0.39 | 0.44 |
| question_open | 0.35 | **0.36** | 0.28 | 0.29 | 0.34 | 0.29 |
| socialact_apology | 0.91 | **0.93** | 0.88 | 0.84 | 0.82 | 0.88 |

Table 3: FScore comparisons of with and without using negation handling for dialogue acts prediction

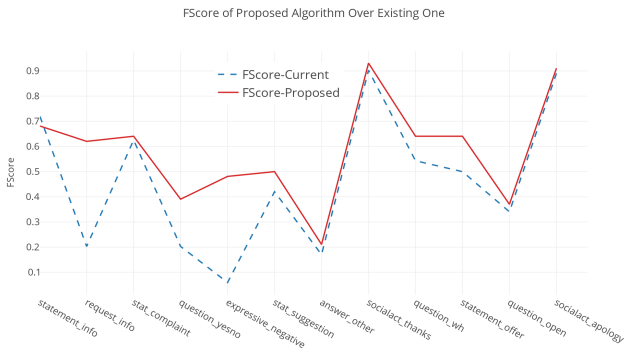

Figure 2: Classification Accuracy Comparison between Shereen et. al [20] and Proposed Dialogue acts prediction algorithm using FScore

## Comparison with Existing work

In this experiment, we compare most recent sequential dialogue acts prediction [20] algorithm in the customer care conversations with our proposed one while handling negation. Note that, the data used in this experiment is different than used in [20]. In this data, we only have instances that have one or more negation cues (Section Data). In Figure 2 we show line plots of FScore for each of the dialogue acts comparing two methods. As we can see the majority of the cases proposed one beats or remain same with the current algorithm. On of the reason, we found out that, in our negation customer care conversation corpus a large portion of data forms one or two length conversations and the sequential approach needs a longer sequence to learn from it.

## CONCLUSION

In this work, we study effectiveness of handling negation in dialogue acts prediction task in twitter based customer care conversation. We create a dialogue acts annotated negation corpus of customer care conversation. We develop several heuristics for handling negation scope and empirically prove that for an informal text like tweets heuristic based approach of negation scope detection performs better.